\documentclass[final,3p,twocolumn]{elsarticle}

\usepackage{amssymb}

\usepackage{amsmath,amssymb,amsfonts}
\usepackage{algorithmic}
\usepackage{graphicx}
\usepackage{textcomp}
\usepackage{xcolor}
\usepackage{url}            
\usepackage{booktabs}       
\usepackage{amsfonts}       
\usepackage{nicefrac}       
\usepackage{microtype}      
\usepackage{hyperref}
\usepackage{url}
\usepackage{dblfloatfix}
\usepackage{wrapfig}
\usepackage{float}
\usepackage[position=bottom]{subfig}
\usepackage{balance}
\usepackage[english]{babel}
\usepackage[utf8]{inputenc}
\usepackage{fancyhdr}

\usepackage{upgreek}

\journal{Building and Environment}

\begin{document}

\begin{frontmatter}



\title{Machine Learning for Smart and Energy-Efficient Buildings}
\author{Hari Prasanna Das\textsuperscript{\textdagger}, Yu-Wen Lin\textsuperscript{*}\textsuperscript{\textdagger}, Utkarsha Agwan\textsuperscript{*}\textsuperscript{\textdagger}, Lucas Spangher\textsuperscript{*}\textsuperscript{\textdagger}, Alex Devonport\textsuperscript{*}\textsuperscript{\textdagger}, Yu Yang\textsuperscript{*}\textsuperscript{\ddag}, J\'an Drgo\v na\textsuperscript{\S}, Adrian Chong\textsuperscript{$\upxi$}, Stefano Schiavon\textsuperscript{$\varpi$}, Costas J. Spanos\textsuperscript{\textdagger}\\
\textsuperscript{\textdagger}Department of Electrical Engineering and Computer Sciences, University of California, Berkeley\\
\textsuperscript{\ddag}School of Automation Science and Engineering, Xi'an Jiaotong University\\
\textsuperscript{\S}Pacific Northwest National Laboratory\\
\textsuperscript{$\upxi$}Department of Building, National University of Singapore\\
\textsuperscript{$\varpi$}Center for Built Environment, University of California, Berkeley\\
\textsuperscript{*}Authors Contributed Equally\\
Corresponding Author: Hari Prasanna Das ({\tt hpdas@berkeley.edu})
}



\begin{abstract}
Energy consumption in buildings, both residential and commercial, accounts for approximately 40\% of all energy usage in the U.S., and similar numbers are being reported from countries around the world. This significant amount of energy is used to maintain a comfortable, secure, and productive environment for the occupants. So, it is crucial that the energy consumption in buildings must be optimized, all the while maintaining satisfactory levels of occupant comfort, health, and safety. Recently, Machine Learning has been proven to be an invaluable tool in deriving important insights from data and optimizing various systems. In this work, we review the ways in which machine learning has been leveraged to make buildings smart and energy-efficient. For the convenience of readers, we provide a brief introduction of several machine learning paradigms and the components and functioning of each smart building system we cover. Finally, we discuss challenges faced while implementing machine learning algorithms in smart buildings and provide future avenues for research at the intersection of smart buildings and machine learning.
\end{abstract}



\begin{keyword}
Smart Buildings \sep Machine Learning \sep Energy Efficiency

\end{keyword}
\end{frontmatter}

\section{Introduction}
\begin{figure*}[!h]
    \centering
    \includegraphics[width=0.9\textwidth]{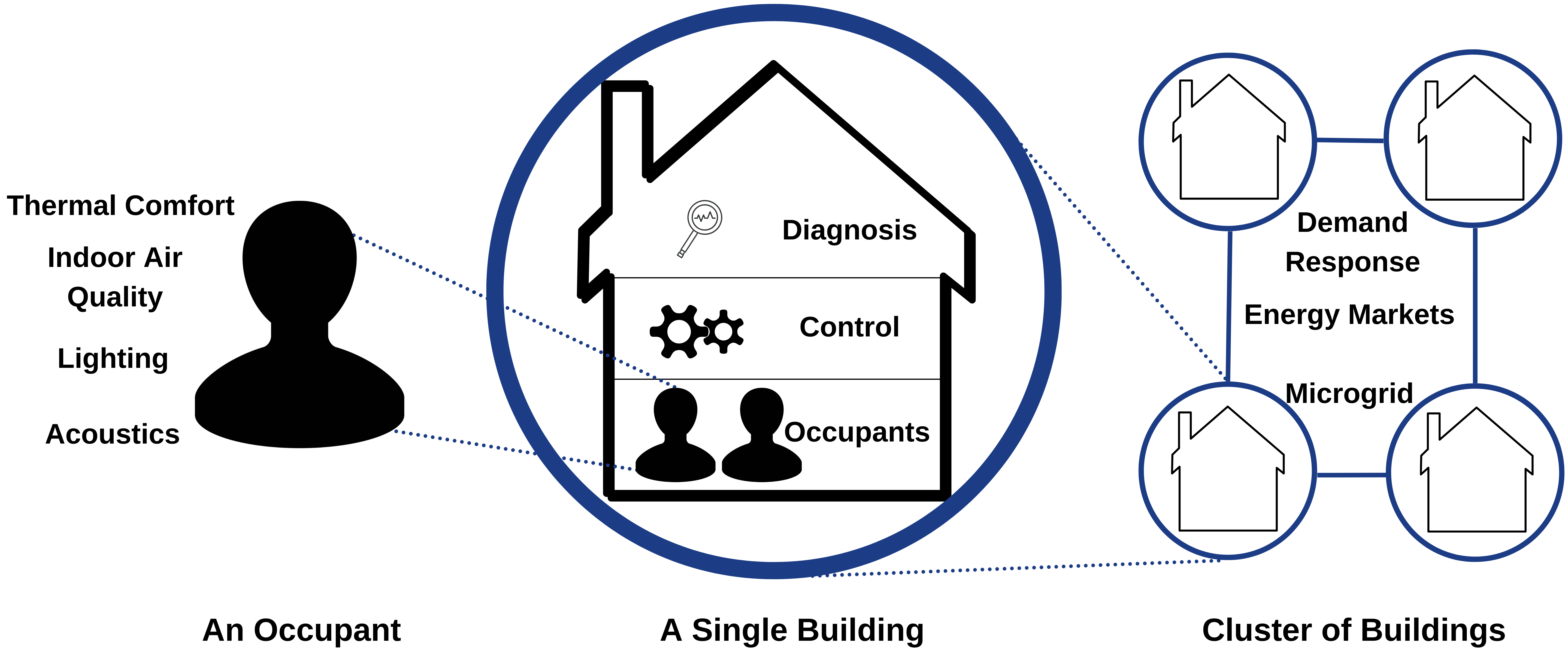}
    \caption{A taxonomy of Smart Buildings ecosystem with building operations illustrated at three levels, at a cluster of buildings level, at a single building level, and at a single occupant level.}
    \label{fig:ecosystem}
\end{figure*}
Buildings and their indoor environment substantially influence our health, well-being, safety, and work and study performance. We spend ~90\% of our time every day in indoor environments. The energy used in buildings to ensure individual safety and comfort is a leading contributor to climate change. It is estimated that in the United States, buildings (both residential and commercial) are responsible for approximately 40\% of primary energy consumption, 73\% of electrical use, and 40\% of greenhouse gas emissions~\cite{DOE}. It is of utmost importance to improve building energy systems to optimize energy usage and thus limit the greenhouse gas emissions contributed by them, while, at the same time, ensuring an occupant-friendly environment to improve well-being and productivity. Energy use reductions while being able to maintain occupant comfort and productivity in buildings can be an environmentally sustainable, equitable, cost-effective, and scalable approach to reducing greenhouse gas emissions.

The smart building ecosystem is illustrated in Fig~\ref{fig:ecosystem}. Occupants constitute the basic building block of the ecosystem. Being the consumer of the facilities that a building provides, occupants necessitate regulation of the building systems to achieve the desired environment. The building comprises of devices and system in place to control and maintain the desired environment for the occupants, along with diagnostic systems to ensure a robust operation. The building operation requires energy, which primarily comes in the form of electricity. The electrical energy is supplied to buildings via a power distribution system, where, with the advent of smart grids, buildings interact and exchange surplus energy with the energy provider and with each other. 

In order to improve energy efficiency and occupant comfort, researchers and industry leaders have attempted to implement intelligent sensing, control, and automation approaches alongside techniques like incentive design and price adjustment to more effectively regulate energy usage. With the growth of internet-of-things (IoT) devices in buildings, the heterogeneity of user-to-device and device-to-device interactions in buildings has become considerable in variety and thus necessitates a system that can adequately account for it. Deriving insights from the vast amount of data in certain scenarios, and from the limited amount of data in other scenarios have also been crucial. Fueled by the above challenges, Machine Learning (ML) as a tool has been implemented in smart buildings, with increasingly important implications. ML constitutes algorithms to process and extract meaningful insights from data and to use it for downstream tasks such as forecasting, prediction, and control. In this work, we explore the status quo of ML applications in smart building systems. In doing so, we touch upon traditional methods already in place, and on how ML-based solutions often match or outperform them.

\section{Machine Learning}

In this section, we briefly cover the fundamentals of various machine learning algorithms and methods that are commonly in practice in smart buildings. Machine Learning algorithms can be commonly grouped under supervised, unsupervised, and reinforcement learning depending on the way a model gets trained on the data-available. We briefly cover the functionalities of the above learning paradigms and various learning methods.

\subsection{Learning Paradigms}
\subsubsection{Supervised Machine Learning}
Supervised learning refers to a class of algorithms that learn using data points with known outcomes/labels.  The model is trained using an appropriate learning algorithm (such as linear regression, random forests, or neural networks) that typically work through some optimization routine to minimize a loss or error function. The model is trained to function by feeding it input data as well as correct annotations, whenever available. If the output of the model is continuous, this process is called regression and if the output is discrete with finite classes, the process is called classification. Supervised learning has been implemented for substantial applications in real-life, such as image classification/segmentation, natural language processing, time-series analysis etc. Since supervised learning required annotated labels as an input, it faces a bottleneck when labels are expensive or require significant labor to obtain.

\subsubsection{Unsupervised Machine Learning}
Unsupervised learning involves models/algorithms that look for patterns in a dataset with no labels and with minimal human supervision. This is in contrast to supervised learning techniques, such as classification or regression, where a model is given a training set of inputs and a set of observations/labels, and must learn a mapping from the inputs to the observations. Some unsupervised learning paradigms include clustering techniques, that aim to group data into similar categories (clusters), dimensionality reduction, which aims to find a low-dimensional subspace that captures most of the variation in the data, and generative modeling, which aims to learn the data distribution and generate synthetic data, etc. 
\subsubsection{Reinforcement Learning}
RL is a type of agent-based machine learning where a complex system is controlled through actions that optimize the system in some manner \cite{sutton2018reinforcement}. The actions ($a_t$) taken based on a probability distribution $p_{\pi}$ at states ($s_t$) at time $t$ seek to optimize the expected sum of rewards$(r)$ based on a policy$(J)$ parameterized by $\theta$; i.e., $J(\theta) = \sum_{s_t,a_t \sim p_{\pi}}[r(s_t, a_t)]$. 

 RL is based on the Markov property of systems, i.e., the evolution of system states is entirely determined by the current state and the subsequent control sequence and independent of the past states and past control inputs. Indeed, RL agents might be thought of as a Markov Decision Processes (MDP) with neural nets serving as function approximators helping to decide which state transitions to attempt. RL is useful in contexts where actions and environments are simple or data is plentiful, with early use cases being optimizing the control of backgammon \cite{tesauro1994td}, the cart-pole problem, and Atari \cite{mnih2013playing}. Recently, much work has been done to extend RL to much broader use cases. 
 
\begin{figure}
    \centering
    \includegraphics[width=\columnwidth]{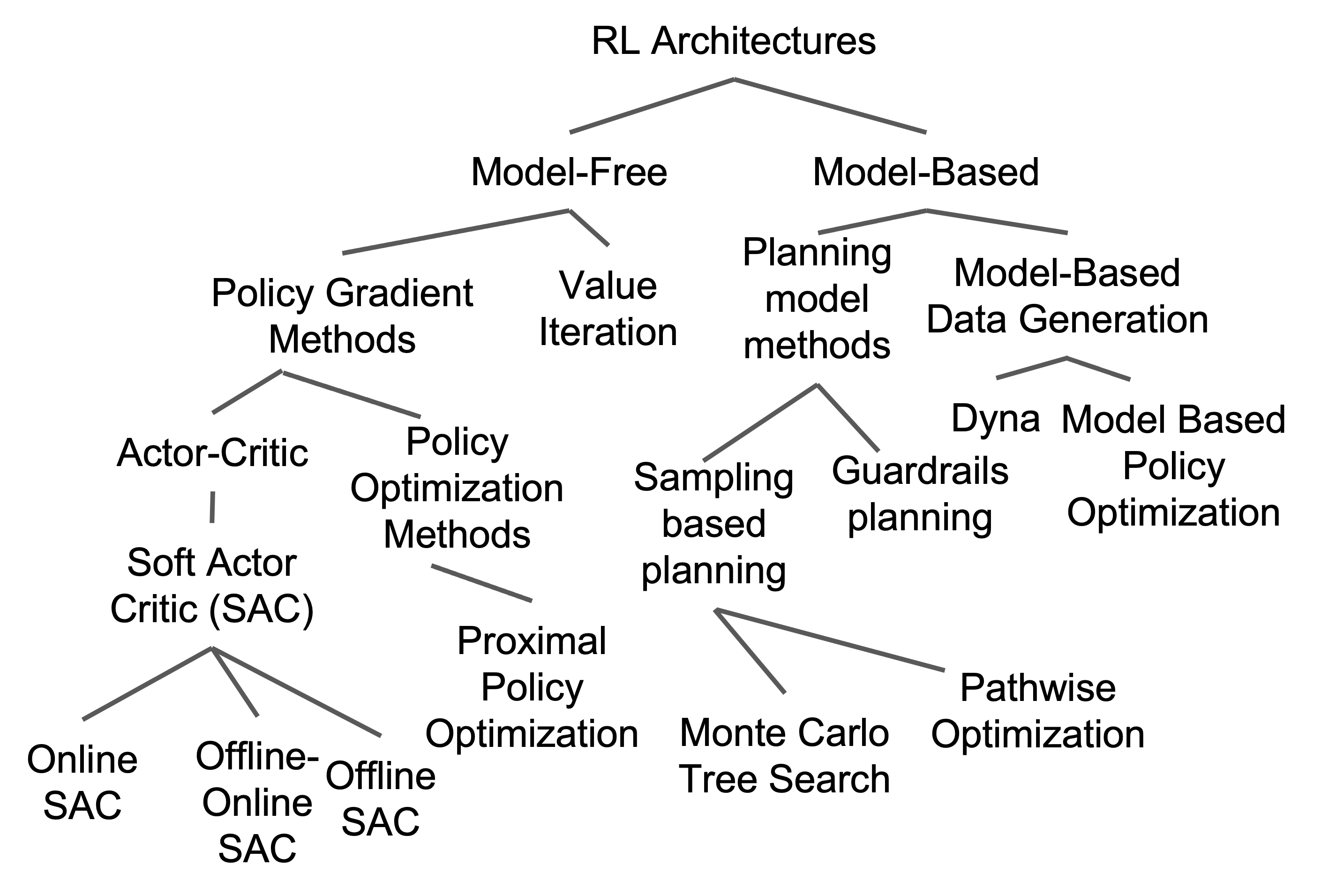}
    \caption{A taxonomy of Reinforcement Learning architectures.}
    \label{fig:RL_tax}
\end{figure}

RL architectures may be grouped in families between which learning differs significantly; a simple one is policy gradient. Here, the controller is made up of a single function approximator that maps actions to states. The model is trained along the gradient of expected reward. One sub-family of policy gradient is the family of Actor-Critic architectures. Here the RL controller is made up of two function approximators; one estimates actions (i.e. the actor),  as in policy gradient, but another estimates the long term value of the actions (i.e. the critic); these estimates contribute to the training of the actor network. An overview of the reinforcement learning architecture taxonomy is provided in Fig \ref{fig:RL_tax}.

We now discuss specific classes of algorithms that are commonly used in the smart buildings domain.

\subsection{Learning Methods}
\subsubsection{Kernel-based Methods}
Kernel algorithms are a class of ML algorithms that are designed for analyzing patterns in data from explicit feature vector representations provided to them. They map the input data into a different space to make the inference of the given data easier by finding hyperplanes in that space that divide the data. Kernel methods can be used for supervised and unsupervised problems. A commonly used kernel-based method for classification is Support Vector Machines (SVM). Similarly, a common method in unsupervised settings is kernel Principal Component Analysis (PCA), that is used for spectral clustering.

\subsubsection{Ensemble Methods}
Ensemble methods combine multiple machine learning models to produce an optimal model. Common examples include tree-based algorithms such as decision tree, bagging (bootstrap aggregating), random forest, and gradient boosting.

\subsubsection{Deep Learning}
Deep learning based methods use artificial neural networks inspired by information processing in biological neurons for applications in supervised, semi-supervised or unsupervised settings. The neural networks are typically designed based on the data they are employed to process, such as convolutional neural networks to analyse image data, recurrent networks to analyse textual, natural language or time-series data etc. 
\subsubsection{Time Series Forecasting Methods}
Time series data comprises of a series of data points indexed by time, such as the stock prices, outdoor temperature or electricity prices. Time series analysis is used to extract statistical information about the data and potentially make predictions about future data points. Wide variety of methods can be used for time series analysis, forecasting, and predictions. Assuming future data has some relation with historically observed data, models such as Moving Average (MA), Auto-Regression (AR), and combinations of these can be used to forecast future data points. Some common use cases are stock price forecasting \cite{pai2005hybrid} and supply chain \cite{aviv2003time}. Recurrent neural networks, and Long Short Term Memory (LSTM) networks can be used for time-series predictions.
\subsubsection{Physics-based Machine Learning}
Machine Learning (ML) models have shown promising results in learning scientific problems that are yet to be well understood or where it’s computationally infeasible with physics based models. However, pure data-driven models often require large datasets, offer limited extrapolation abilities, and are unable to provide physically consistent results. On the other hand, pure physics-based models are often complex, include large amount of uncertain parameters, and may suffer from burdens of computational cost. As a result, research communities have begun to explore the combination of physics domain knowledge and ML models. The hybrid models may compensate the weaknesses of pure physics-based models or pure data-driven models, and provide potential for achieving generalizability and explainability our of the models. 
Coupling physics knowledge with ML has been used in several areas, including medicine \cite{PINNcardiac2020, KISSAS2020_PINNcardio}, fluid mechanics \cite{Cai2022_PINNfluid}, power systems \cite{Misyris2020_PINNpower}, manufacturing \cite{QI2019_PINNmanu} etc. It has shown potentials in better prediction accuracy using a smaller set of sample data and generalizability for out-of-sample scenarios. 

\subsubsection{Learning, Dynamics, and Control}
Control theory has at times anticipated the need to learn about the system from data: this is collected in the disciplines of system identification~\cite{ljung1998system} and adaptive control~\cite{ioannou2006adaptive}, which have similarities to supervised learning~\cite{ljung2020deep} and reinforcement learning~\cite{matni2019self} respectively. While some system identification is implicit in all control approaches, as a part of how the model is formed, adaptive control never rose to prominence as reinforcement learning has.

From the reinforcement learning side, several approaches to model-based reinforcement learning use control-theoretic methods. Essentially, the problem is split into two parts: a dynamical model is learned from data, and the model is used to design a controller, with variations on model-predictive control being a popular choice. From the control theory side, learned models are incorporated into modern control synthesis techniques, and the standard methods of proving safety are adjusted to incorporate learned models. A learning-based model that has proved to mesh particularly well with this approach are Gaussian process models, which have experienced a renaissance among learning-based controls researchers~\cite{umlauft2017learning, wang2018safe, akametalu2014reachability, devonport2020bayesian, devonport2021data}.
Another point of contact is the use of statistical guarantees for controllers based on learned models. In reinforcement learning, statistical guarantees of correctness such as regret bounds and probably approximately correct (PAC) bounds are used to ensure that optimal behavior is attained with high probability. By combining these statistical guarantees with control-theoretic techniques, they can also be used to provide guarantees of safety with high probability~\cite{devonport2021data, devonport2020estimating, 9682860}. Since many learned models are statistical in nature, such a probabilistic bound is often the best possible without overestimating the accuracy of the model, which has also become popular in recent years~\cite{matni2019self}.

\section{Applications of Machine Learning in Smart Buildings} 
\begin{figure*}[!h]
    \centering
    \includegraphics[width=\textwidth]{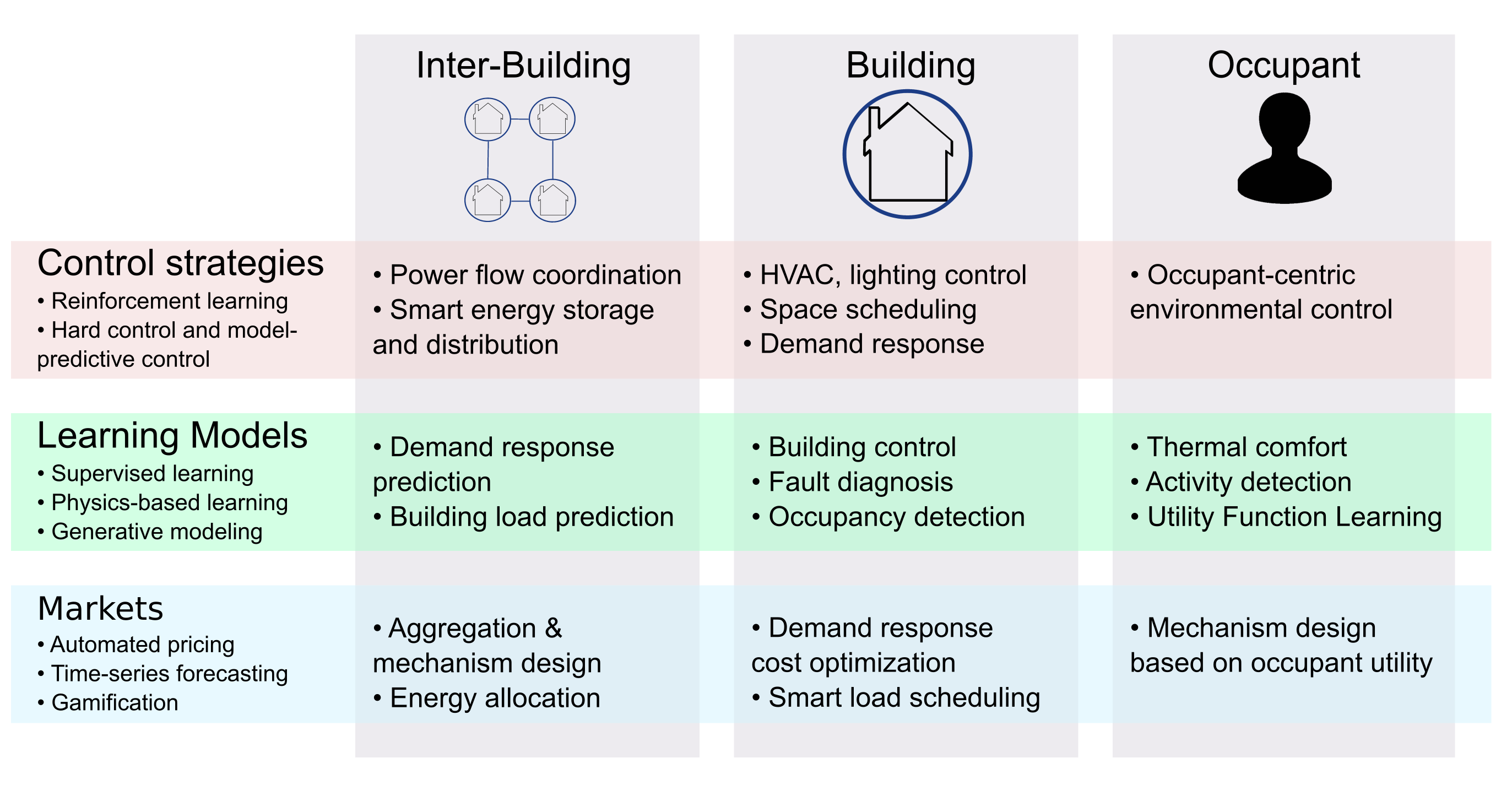}
    \caption{Illustration of various applications where machine learning methods can be deployed in smart buildings, grouped at the cluster of buildings-level, the building-level, and the occupant-level.}
    \label{fig:basketweave}
\end{figure*}

We organize the ML applications in smart buildings into different building sub-systems and other desiderata. A summary of ML applications at the level of occupants, single building and a cluster of buildings is provided in Fig.~\ref{fig:basketweave}.  

\subsection{Thermal Comfort} \label{sec:thermal-comfort}
The American Society of Heating, Refrigerating and Air-Conditioning Engineers (ASHRAE) defines thermal comfort as \textquotedblleft the condition of the mind that express satisfaction with the thermal environment and is assessed by subjective evaluation\textquotedblright~\cite{standard2020standard}. Humans spend more than 90\% of their days within a built environment, where their health, well-being, performance and energy consumption are linked to thermal comfort. But, studies show that only 40\% of commercial building occupants are satisfied with their thermal environment~\cite{graham2021lessons}. Ensuring occupants' thermal comfort involves understanding of the parameters that affect it, developing and using predictive models to predict thermal comfort, and controlling HVAC systems to achieve occupant satisfaction. Machine learning has been applied to all of these sectors. In this section, we mainly cover thermal comfort understanding and prediction. In Section~\ref{sec:control}, we cover the control of the thermal environment in buildings. Thermal comfort understanding and prediction for occupants in a building have been studied extensively in past.  The prediction models can be broadly divided based on the features used to develop the model, as traditional models like Predicted Mean Vote (PMV) and the adaptive thermal comfort models and the more innovative personal comfort models.

The most widely used thermal comfort model is the PMV, developed by Fanger in 1970~\cite{fanger1970thermal} based on a set of experiments in controlled climate chambers. Fanger established a mathematical formula with six input parameters: air temperature, mean radiant temperature, relative humidity, air velocity, clothing insulation and metabolic rate. 
The PMV model adopts a 7-point scale for its thermal sensation output, with values ranging from -3 to +3, and indicating cold, cool, slightly cool, neutral, slightly warm, warm, hot, respectively. To gauge the level of dissatisfaction of individuals in a space where PMV has been computed, Fanger proposed a Predicted Percentage of Dissatisfied (PPD) model~\cite{fanger1970thermal}, that establishes a quantitative prediction of the percentage of thermally dissatisfied occupants. To comply with ASHRAE 55 standards, the recommended range for PMV on the 7-point scale is between -0.5 and 0.5. The PMV model is designed with a uniform and steady-state conditioned environment and does not explicitly consider local discomfort, the non-uniformity of the space and dynamic thermal conditions~\cite{zhao2021thermal}. It has also been developed with data from controlled climate chamber experiments, which are not necessarily transferable to the varying thermal environments that exist in buildings around the world~\cite{cheung2019analysis}. Experiments in real buildings showed that the PMV has a low prediction accuracy (around 33\%), whereas the PPD is unreliable as a metric, hence it is not advisable to be used~\cite{cheung2019analysis}.

With the booming development of Machine Learning (ML) techniques and their versatile applications, researchers have attempted to develop data-driven approaches using ML to predict thermal comfort. Often, the Fanger's features, alone or with additional relevant real and synthetic features, are fed into a data-driven model to learn the connections between the features and the thermal preference labels~\cite{liu2019personal,kim2018personal}.
The model is later leveraged to predict the same given raw features. Because the models are trained to learn from the data itself, and not from a rule that was established using prior experiments, they perform better as compared to PMV. 

Kernel-based approaches have been the widely-used methods for thermal sensation/preference prediction. The list of kernel-based methods popular for thermal comfort prediction are Support Vector Machine (SVM)~\cite{zhou2020data,chai2020using,alsaleem2020iot,chaudhuri2018thermal,liu2020machine}, K-Nearest Neighbors (KNN)~\cite{pigliautile2020assessing,cheung2022impacts,lu2019data,lee2021physiological,liu2019personal}, and Ensemble Learning algorithms, such as Random Forest (RF)~\cite{liu2019personal,kim2018personal} and AdaBoost (Ab). Recently, neural network such as feed-forward networks~\cite{das2021unsupervised,lu2019data,zhai2017balancing}, and time-series based networks~\cite{chennapragada2022time} have proven to be capable of surpassing the state-of-the-art kernel-based models in thermal comfort prediction.

Traditional thermal comfort models, as described above, are developed based on aggregated data from a large population. They were designed to predict the average thermal comfort of the entire population rather than an individual. 
Personal thermal comfort models, which is a new approach to thermal comfort modeling that predicts individuals' thermal comfort responses instead of the average response of a large population, can be applied to any building thermal control system~\cite{kim2018personal}. Personal thermal comfort can adapt to the available input variables, such as environmental variables~\cite{cheung2017longitudinal}, occupant behaviors~\cite{kim2018personal1} and physiological signals~\cite{liu2019personal}. Several experiments have been conducted over the years for population groups varying in terms of environmental and personal conditions. ML algorithms ranging from kernel-based methods to neural network based methods have been proposed as the predictor. For instance,~\cite{liu2019personal} conducted an experiment to collect physiological signals (skin temperature, heart rate) of 14 subjects (6 female and 8 male adults) and environmental parameters (air temperature, relative humidity) for 2–4 weeks (at least 20 h per day). They developed ML based predictors of thermal preference for each individual, using synthetic features derived from the above raw features (e.g. mean heart rate over last 5-min from the survey response time) as the input with Random Forest being the best predictor. A data-driven method with indoor environment was applied to classify occupant's personal thermal preference with temperature and humidity sensors~\cite{laftchiev2016iot}. Authors in ~\cite{sim2018wearable} developed personal thermal sensation models based on watch-type sweat rate sensors. Among all the ML methods, kernel based algorithms are still the most commonly used ones. However, other ML approaches are also becoming popular in modeling the complex interactions that exist between the features without much feature engineering, e.g. time-series prediction ~\cite{somu2021hybrid}, artificial neural networks~\cite{das2021unsupervised}, etc. Better approaches for modeling tabular data in smart buildings, with a focus on thermal comfort datasets are provided in ~\cite{das2022improved}.

Although several studies show high prediction accuracy in predicting thermal comfort with various ML methods, there are still several challenges ahead. A common challenge in designing ML based thermal comfort predictors is the issue of class-imbalance in data. Almost all of the thermal comfort datasets~\cite{livcina2018development,liu2019personal} are inherently class-imbalanced, i.e. they have more data belonging to \textquotedblleft Prefer No-Change\textquotedblright than \textquotedblleft Prefer Warmer\textquotedblright and \textquotedblleft Prefer Cooler\textquotedblright thermal preference classes. Researchers have tackled this issue by weighed loss function for ML models. A related challenge is the overall lack of sufficient amount of data~\cite{liu2019personal}. Collecting large amounts of data as required by ML models from humans via real-world experiments is cumbersome and resource-intensive. Generating synthetic data to augment the training dataset is one of the approaches that has been proposed for tackling with the above challenges. The synthetic data generation can be done using classical methods such as SMOTE~\cite{chawla2002smote,quintana2020balancing} or using advanced neural network-based generative models~\cite{quintana2020balancing,yoshikawa2021data,das2021conditional,das2022conditional}.

Another challenge is domain discrepancy. Thermal comfort, as per PMV, is dependent upon 6 major parameters as described in the beginning of the section or, as per the adaptive thermal comfort model, is dependent on the outside temperature. However, it also varies from person to person, across climatic regions and economic conditions. A literature review of personal comfort models concluded that there is a lack of diversity in terms of building types, climates zone and participants that are considered in existing thermal comfort studies~\cite{martins2022systematic}. Under such domain discrepancy, using models developed in one environment in another target environment may lead to low accuracy or misleading predictions. Thermal comfort modeling in the target domain independently requires a large amount of self-reported thermal comfort data. To deal with the data/label insufficiency challenge, domain adaptation methods have been proposed to adapt ML models from one domain to another. ~\cite{das2021unsupervised} propose a transfer learning framework, using Adversarial Domain Adaptation (ADA) to develop personal thermal comfort predictors for target occupants in an unsupervised manner. ~\cite{gao2021transfer} propose a transfer learning-based multilayer perceptron model for domain adaptation across cities. 

Summarizing, machine learning models have proven to increase thermal comfort predictions accuracy and they allow a wider and more flexible range of input parameters. They can adapt to the available data streams, and they improve their performance over time. There are several future research in this direction that we enlist in Section~\ref{sec:future_works}.
\subsection{Occupancy and Activity Sensing} 
Occupancy and activity sensing are key aspects for observability of a human-in-the-loop building control system. Traditionally, building operation methods that include occupancy as one of their parameters, such as starting heating/cooling from early morning till
late in the evening during weekdays assuming maximum
occupancy during working hours often have static schedules set for the occupancy, which is far from realistic. Also, How much a building will be occupied depends on several other factors, such as weather, building type, and holiday schedule. Such static policies may lead to a significant waste in energy consumption, because the heating/cooling and ventilation levels are set with no regard for the actual occupancy level. Activity sensing also helps to provide personalized, context-aware services in buildings, thus enhancing overall satisfaction in buildings while creating a safety net for adverse events such as falls in elderly homes~\cite{petrosanu2019review}.

Occupancy sensing can be performed using both intrusive and non-intrusive methods. Intrusive methods require the occupants to carry an electronic device whose signature is followed by a central server to infer occupancy/positioning~\cite{lee2013occupancy,zou2018unsupervised,zou2016bluedetect,zou2018winlight,filippoupolitis2016bluetooth}. However, requiring occupants to constantly carry a device is not reliable. This problem gets magnified for the case of elderly population. Hence, non-intrusive methods for occupancy sensing are getting popular. In non-intrusive method, occupancy/activity is detected from data collected from various sensing modalities, and some ground truth during controlled experiments. Machine-learning based methods are very effective for non-intrusive occupancy sensing, allowing for the data-driven construction of a model to map sensor data into an estimate of occupancy.

Non-intrusive methods for occupancy sensing can be divided according to the sensing modalities they employ. A common choice is video feeds from cameras installed in rooms. Regular cameras have not found significant usage because of their privacy issue, rather infrared sensors or thermal cameras have been the choice for occupancy detection. For example, authors in ~\cite{kraft2021low} use an U-Net like convolutional neural network on thermal images to infer occupancy. Other works employing similar machine learning methods on depth cameras are ~\cite{zhao2019occupancy,diraco2015people,brackney2012design}. But, in general, cameras have other issues such as poor illumination conditions and occlusion. A recent body of work focuses on occupancy and activity detection from WiFi signals~\cite{zou2019machine}, because of its ubiquitous presence, and better privacy guarantees. Authors in ~\cite{zou2018device,zou2017freedetector} use Channel State Information (CSI) data collected from WiFi sensors (a transmitter and a receiver) and measuring the shape similarity between adjacent time series CSI curves to infer the occupancy. They improve the detection mechanism in ~\cite{zou2019consensus} by using convolutional neural networks on the CSI heatmaps to detect human gestures. Another modality that is used to detect occupancy is CO$_2$ data in a room. A number of works~\cite{zuraimi2017predicting,arendt2018room,rahman2017occupancy,wei2022deep} employ machine learning methods to map the CO$_2$ concentration and occupancy. Finally, works such as ~\cite{zou2019wifi} propose sensor fusion, where data from multiple sensing modalities, i.e. RGB camera, and WiFi are used in tandem to come up with a robust activity detection mechanism. 

To summarize, machine learning methods have been proposed to map the relation between data from sensors, and occupancy/activity. Once trained, these models can then be deployed to predict the occupancy which then gets fed to building control mechanisms~\cite{esrafilian2021occupancy}.
\subsection{Building Design and Modeling}
\begin{figure}[t]
    \centering
    \includegraphics[width=0.48\textwidth]{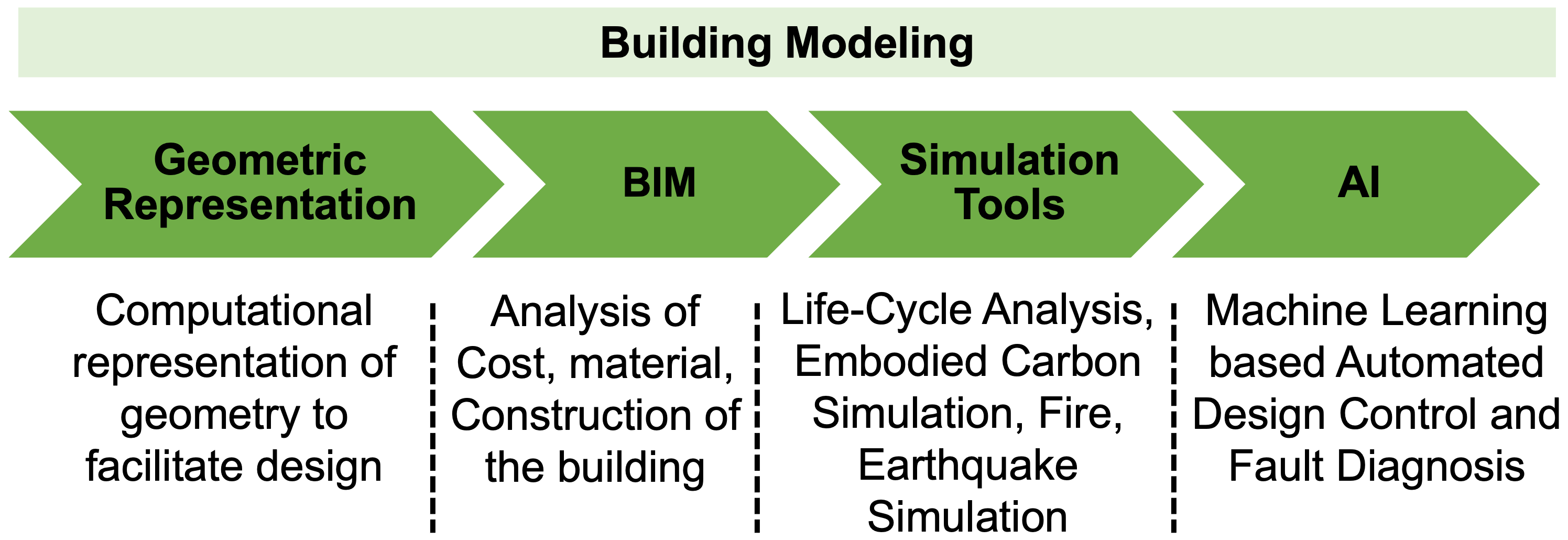}
    \caption{Evolution of building modeling.}
    \label{fig:model_evo}
\end{figure}

Building Information Modeling (BIM) was introduced in the 1970s when Eastman \cite{eastman1975use} proposed the idea of migrating from hand-drawn building design towards automation and digitization.
Early building modeling focused on the computational representation of geometry, which is also known as 3D modeling.
Later, a set of new tools focusing on the building performances were developed, among them Building Design Advisor (BDA) \cite{PAPAMICHAEL1997BDA}, Ecotect \cite{roberts2001ecotect}, EnergyPlus \cite{crawley2001energyplus}, ESP-r \cite{espr}, DEST \cite{yan2008dest}, and Modelica \cite{fritzson1998modelica}. They focused on thermal, energy, lighting and air quality aspects. They provided feedback and ``what-if" analyses.
With the emergence of IoT, data has become more easily accessible, which has helped support building simulation tools and allowed development of machine learning models.
The development of building models over the past decades is summarized in Figure \ref{fig:model_evo}.

The goal of building modeling is to support the design, construction, and operation of buildings.
Early building modeling was mostly used during the design phase, and was not usually maintained or updated after the building was constructed.
In recent years, the digital twins (DT) concept \cite{Grieves2017} has emerged for physical systems where a virtual ``twin" is developed that mirror the actual physical system. 
DT can monitor a building's condition in real-time and plays an important role in control applications, fault diagnosis and prognosis. 
The following sections describe the intersection between machine learning and building modeling.
It is separated into five parts: models, design automation, digital twins, applications, and challenges.

\subsubsection{Models}
\begin{figure}[t]
    \centering
    \includegraphics[width=0.48\textwidth]{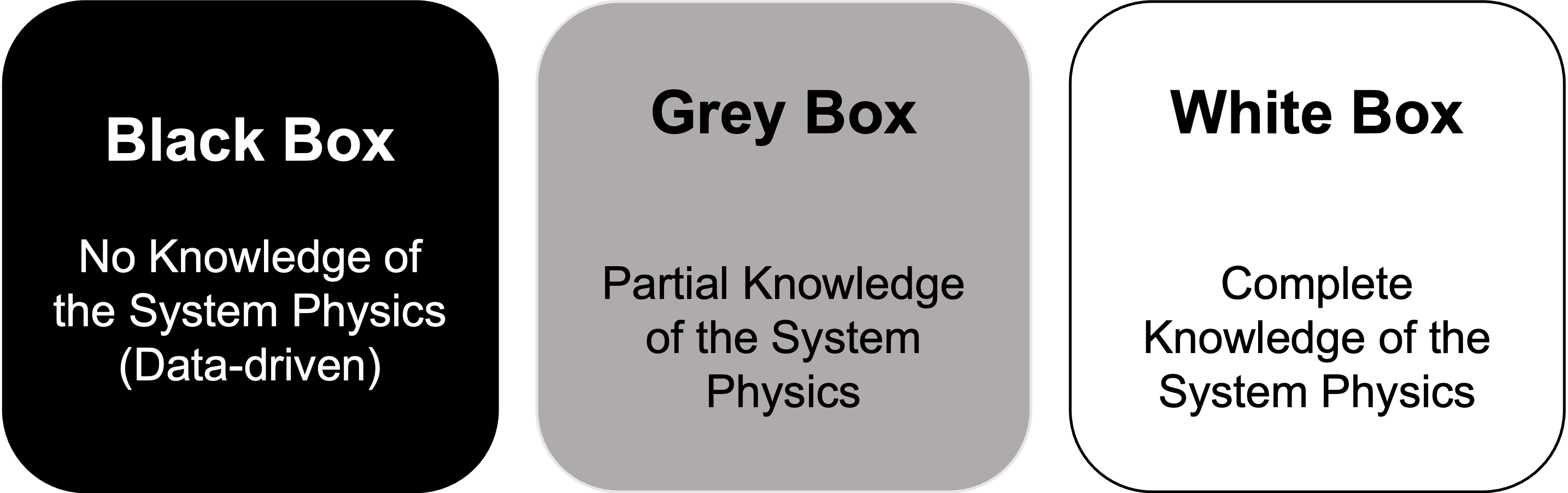}
    \caption{Three types of building models}
    \label{fig:design}
\end{figure}
There are three types of building models: white box, black box, and grey box, as illustrated in Figure \ref{fig:design}. 
\paragraph{White box model}
The ``white box'' or ``first principles'' approach models the building system with detailed physics-based equations.
Building components and subsystems are modeled in a detailed manner to predict their behaviors. 
Common Building Performance Simulation (BPS) tools for physics-based modeling are Modelica \cite{fritzson1998modelica}, EnergyPlus \cite{crawley2001energyplus}, TRNSYS \cite{trnsys}, and ESP-r \cite{espr}. Though a white box model can capture the building dynamics well, it requires a lot of information from the physical building and may suffer from high uncertainties.

\paragraph{Black box model}
A black box model, also known as a data-driven or machine learning model, is a modeling approach that constructs a model directly from data without knowledge of the system physics.
With the emergence of Internet of Things (IoT), sensor measurements are continuously collected at various locations in a building to reflect current operations, which provides an opportunity to build data-driven models.
For example, building load prediction using black box models is an important aspect of improving building energy performance. 
Its applications include control optimization, fault diagnosis, and demand side management \cite{zhang2021review}. 
The common machine learning methodologies for building load prediction are ANN, SVM, Gaussian-based regressions, and clustering \cite{seyedzadeh2018machine}. 
The drawbacks of black box models include the need for a large amount of data and the fact that the accuracy of the model is highly dependent on the data quality. Another drawback is their fragility, which occurs when they give results that are not physically possible and may not perform reliably in conditions that have not been seen before. In addition, black box models lack interpretability and may have a high computational cost. 

\paragraph{Grey box model}
Grey box modeling is a modeling approach that simplifies the physical equations to simulate the behavior of a building and it combines them with machine learning techniques.
By doing that, it aims to reduce the system complexity while maintaining prediction accuracy. 
An example of such a model is using thermal networks to model buildings \cite{bacher2011identifying}, which is a common approach in building modeling. 
Thermal networks model dynamics of a building through resistor and capacitor (RC) circuits, and the parameters are identified with measurements from the real system. 
Studies show that the grey box model works well in building control applications \cite{oldewurtel2012mpc, vsiroky2011experimental} and grid integration.
However, the design of the thermal network is still ambiguous.
The best selection of the number of resistors and capacitors for the model is unclear. 
Bacher et al. \cite{bacher2011identifying} analyzed the model selections and found that the 3R3C model is the minimal best fit model for a $120m^2$ building located in Denmark.
However, it is inconclusive if all buildings of similar size and use can be modeled with 3R3C with the same accuracy.

Lin et al. \cite{Lin2021hybriddt} build a building model based on information available to minimize the number of uncertain parameters by combining physics-based equations with a neural network.
Other researchers \cite{daw2017physics,robinson2022physics} explored physics-based machine learning for modeling lake temperature and other dynamical systems.
However, there are limited studies in physics-based ML for building modeling.

The definition of grey box models is still ambiguous, where it can refer to a Resistor-Capacitor Network or a hybrid model that utilizes both first principle equations and data-driven methods. 
While grey box models have shown success in several applications, they are not systematically defined or thoroughly analyzed.

\subsubsection{Design Automation} 
Buildings are complex systems. Among their objectives is to provide comfortable and safe conditions to people. Building performance is affected by people's behavior \cite{HONG2017518}. Developing a model of a building with these complexities is often time-consuming and costly. 
Though white box models are widely used in simulation-based studies, grey and black box model are more commonly used in experimental studies of real buildings~\cite{zhan2021data}. 
As a result, building design automation is an active area of research to help in developing fast and accurate building models. 

Currently, automated design of buildings mostly involves the development of 3D building models. Most of the 3D models are built with Revit/Autocad based on site measurements or estimations during the design phase.
Sometimes, researchers collect building measurements through LiDAR or airborne stereo imagery to construct 3D building models \cite{haala2010update}, which can automatically be imported to BIM or building simulation tools.
However, this approach only captures the building at a geometrical level and cannot represent the operational level of buildings.
Jia et. al. \cite{jia2018design} proposed a platform-based design to reduce redundancy in hardware and software usage in building applications.
The design performance is optimized by exploring the design space. 

The emergence of building design libraries such as Modelica Building Library \cite{Wetter_modelicabuildinglibrary} opens up opportunities for platform-based design automation of buildings. 
However, automated building modeling that captures building dynamics is underdeveloped, and most modeling methods still require human intervention. 
\subsubsection{Digital Twins}

The (AIAA) and (AIA) \cite{aiaa2021digital} recently published a position paper that defines a digital twin as ``\textit{A set of virtual information constructs that mimics the structure, context and behavior of an individual / unique physical asset, or a group of physical assets, is dynamically updated with data from its physical twin throughout its life cycle and informs decisions that realize value.}'' The same definition can be applied to buildings, specifically how a physics-based model evolves over the course of its lifecycle as information or data about the building becomes available, i.e., as the building goes from the design phase (as designed) to the construction and commissioning phase (as built) to the post-occupancy phase (“as operated”). For buildings, forward building performance simulation (BPS) tools such as EnergyPlus \cite{crawley2001energyplus}, TRNSYS \cite{trnsys}, ESP-r \cite{espr}, and Modelica \cite{fritzson1998modelica} are usually used to estimate the energy use of the building and its subsystems. Over the past decade, machine learning has been increasingly applied to BPS, fueled in part by the emergence of the Internet of Things (IoT) and the need for computationally tractable predictions. 

Machine learning has been used during the design stage to augment generative design and parametric simulations. Deep generative algorithms such as Generative Adversarial Networks (GANs) \cite{huang2018architectural, nauata2020house} have been proposed for generating diverse but realistic architectural floorplans that are known to be a time-consuming iterative process. The automated generation of architectural floorplans can be coupled with BPS tools to systematically explore architectural layouts that optimize building energy efficiency \cite{gan2019simulation}. 

Metamodeling, defined as the practice of using a model to describe another model as an instance~\cite{allemang2011expert}, is another aspect where machine learning has been extensively applied to BPS throughout the building lifecycle. Given the complex interaction between different building systems and sub-systems, design optimization during early design typically requires exploring high-dimensional design space. Consequently, machine learning has been used to create metamodels that can be used for optimization and uncertainty analysis \cite{eisenhower2012methodology, bre2020efficient}. Although most studies predict energy consumption and thermal comfort, metamodels have also been proposed for the emulation outputs such as natural ventilation and daylighting to support the modeling of passive design strategies \cite{chen2017developing}. A comparison of different metamodeling techniques can be found in \cite{ostergaard2018comparison}.

Moving from design to post-occupancy, model calibration is often undertaken to improve the model's credibility by reducing the discrepancies between simulation predictions and actual observations. Metamodeling using machine learning is often used to emulate simulation predictions to alleviate the high computation costs \cite{coakley2014review}. Particularly, metamodeling is almost always applied in Bayesian calibration that is computationally intractable because of the need to perform many model evaluations \cite{chong2021calibrating}. More recently, Bayesian optimization and meta-learning through deep probabilistic neural networks have been proposed to reduce the number of simulations required for model calibration significantly, avoiding the need for metamodels altogether \cite{chakrabarty2021scalable,zhan2022calibrating}.

BPS has also been applied to enhance the use of deep reinforcement learning (DRL) based building controls. For instance, BPS can be used to develop building test cases for the comparison and benchmarking of DRL algorithms \cite{blum2021building}. The deployment of DRL in real-world building controls can be challenging because of the need for large amounts of training data before achieving acceptable performance \cite{botvinick2019reinforcement}. Consequently, the exploration during online training can lead to undesirable building operations when the algorithm performs sub-optimally. To enhance the practical applications of DRL, BPS has been used as an emulator to pre-train a DRL agent offline. For instance, Zhang et al. \cite{zhang2019whole} used an automatically calibrated EnergyPlus model for the offline training of an A3C (Asynchronous Advantage Actor Critic) algorithm that is subsequently deployed in a radiant heating system. It was also shown that a DRL trained offline could achieve similar near-optimal results as model predictive control \cite{brandi2022comparison}. However, pre-training a DRL agent using a model contradicts the advantages of DRL being ``model-free'' because BPS models can be time-consuming to develop. Therefore, further research could usefully explore the use of simpler and a lower level of detail models for the offline training of DRL agents for building controls.

\subsubsection{Applications: Design, Control and Fault Diagnosis}
The main applications for building modeling are facilitating design choice, testing control algorithms, and detecting faults in a system. 
BPS is used in the design phase of buildings to provide feedback in energy consumption, and greenhouse gas emissions and to guide design choices \cite{hensen2012building, clarke2007energy}. 
Additionally, the development of a robust control system often requires a detailed understanding of system dynamics. 
Building modeling can provide dynamic feedback to test various control algorithms, which will be discussed in more detail in the next section. 
Fault detection also plays a crucial part in reducing maintenance costs and increasing the energy efficiency of building operations \cite{dong2014bim}. 
However, faults in an actual building don't occur frequently, and it is hard to collect fault data for analysis. 
Data sets for fault diagnosis are usually created through testbed experiments or simulations.
BPS can provide ``what-if" analysis to assist in fault modeling and creating meaningful datasets. 

\subsubsection{Challenges}
There are two main challenges in building modeling.
First, there is a gap between building models during the design and the operation phases. 
Moreover, no standard connections are established between BIM and BPS tools, which results in unavoidable human effort to recreate the model for operation after the design phase. 
Though some established geometrical models can be exported to BPS software, there are limitations in software compatibility and set up requirements. 
Second, there is no systematic approach to model selection. 
Limited studies have explored whether a certain type of model (white, black, or grey box model) is better for a specific application. 
Developing a level of abstraction of building modeling may benefit the development of a better model selection process for various applications. We anticipate that techniques from machine learning, such as neural architecture search and physics-based ML can be used here to improve building modeling.
\subsection{Building  Control}\label{sec:control}
A critical application of a building model is the design of control systems for the building subsystems. These are typically \textquotedblleft local models\textquotedblright that capture only  aspects of the building that are pertinent to low-level control, such as a physical model heat and ventilation in a collection of rooms near a target temperature. However, a large-scale model is still important for controller design, as it is necessary to understand how the subsystems of a building interact. For the purposes of this review, the most important subsystem in terms of complexity of control and size of energy demand is arguably the Heating, Ventilation and Air Conditioning (HVAC) system, so we will focus our attention there. However, we do want to note that HVAC is by no means the \textit{only} controls problems that has received focus; other building control systems enjoy active work and engaged communities,  including lighting \cite{ul2014review, panjaitan2011lighting, wagiman2019review, lee1994design}, window control \cite{yoon2020optimization, psomas2017ventilative}, electric vehicle charging \cite{turker2018optimal, al2019review, kontou2017cost}, and domestic water heating \cite{wanjiru2017optimal, starke2020real, passenberg2016optimal}. 

Controlling the HVAC system of a building can be, and for many decades has been accomplished by classical control techniques. Here, the control objective simply being to keep temperatures somewhat close to a setpoint or within a deadband. Two simple controller types have proven to achieve this goal for HVAC systems: (1) threshold-based \textquotedblleft bang-bang\textquotedblright controllers (2) and PID controllers. (1) Bang-bang control applies the greatest allowable heating or cooling when the measured temperature leaves a certain band around the target temperature, but otherwise remains inactive. (2) PID controllers use a linear feedback strategy, which decides a control action using a linear function of previous measurements. PID denotes a \textquotedblleft Proportional-Integral-Derivative\textquotedblright  function formed as the sum of the current measurement (the proportional component) and approximations of the time derivative and integral of the temperature. These simple control policies even achieve optimality in certain cases: for example, the bang-bang strategies are optimal for minimizing the total time that the HVAC system is active, and PID controllers can be tuned to minimize the sensitivity of the controller to exogenous disturbances. However, if we ask for a strategy that is optimal with respect to delivering thermal comfort, or to account for energy load profiles throughout the day, predictively pre-heat and cool, or try to minimize other metrics such as carbon emissions, these simple control strategies no longer suffice.

The optimal operation of HVAC systems with respect to cost, emissions, comfort, or a combination therein, is a multi-step optimization problem due to the thermal capacity of buildings and the nonlinear operation of the HVAC system. 
Specifically, the dynamics of building thermal behavior is a slow-moving process and subject to time delays. This enables the precooling or preheating of buildings (e.g., storing the cooling energy in building mass).
Besides, the operation efficiency of HVAC system is nonlinear w.r.t. supplied air. For example, the amount of energy required to maintain a constant temperature in a zone is a product of supplied zone air flow rates and the difference of the indoor temperature and the supplied temperature. The fan power of air handling unit (AHU) is cubic with respect to the total supplied air.  In addition, the operation of HVAC system suffers from time-varying disturbances, such as  the  outdoor weather variations (i.e., radiation, temperature, humidity, etc.) and the occupancy changes. 

HVAC system operation has raised extensive concerns in recent decades due to the continued increase of energy consumption of buildings and the high priority to save building energy consumption. Fortunately, the related work is plentiful and rapid progress has been achieved in this area.  Existing works can be broadly categorized by model architecture into: (1) \emph{Model Predictive Control} (MPC) and (2) \emph{Reinforcement Learning} (RL). On top of that, the combination of ML with MPC is developed with the objective to explore the benefits of the classic MPC controllers and the powerful approximation and characterization capabilities of ML tools. 

In what follows, we review the existing works in the three categories.

\subsubsection{Model Predictive Control (MPC)}

Model predictive control takes a large step towards improving these aims by incorporating optimization directly into the control policy. We wish to frame briefly our section on MPC: although it would not be considered a classical case of ML, we find it helpful to thoroughly embed the motivations behind its development and active use into our understanding of the learning-based methods to follow. Thus, we will spend significant time describing it. 

The key idea of MPC is to predict the the value of the objective function in the future using an auxiliary dynamical model of the building's behavior, and to select a control action that optimizes the predicted reward.
Generally, if the dynamics model is $f$, the action  $u$, the state $x$, the objective $y$, and the current timestep $t$, then MPC solves:

\begin{equation}
    \label{eq:mpc}
    \begin{aligned}
     \underset{u}{\text{argmin}} & \sum_{\tau=t}^{t+T} -y(x, u) \\
     \text{subject to: }& x(\tau+1) = f(x(\tau), u(\tau));  \tau= t, \ldots, T-1\\
     & x(t+T) = 0;  \tau= t, \ldots, T-1\\
     & x(\tau) \in \mathcal{X};  \tau= t, \ldots, T-1\\
     & u(\tau) \in \mathcal{U};  \tau= t, \ldots, T-1\\
    \end{aligned}
\end{equation}

The condition $x(\tau) \mathcal{X}$ is a state space constraint, which can be used to ensure safety by attempting to keep the state vector in a region of the state space that is known \emph{a priori} not to be dangerous.

For a given state $x$,~\eqref{eq:mpc} computes a sequence of $T$ control actions in order to minimize the predicted reward. While the building control policy \emph{could} use all $T$ actions, so that~\eqref{eq:mpc} need only be solved every $T$ time steps, to do so would force the controller to ignore what is happening in the outside world for those time steps. 
Since the control model used in~\eqref{eq:mpc} can never be fully accurate, and since dynamical prediction errors compound very quickly, what happens during those those time steps is likely very different from what the controller predicts in~\eqref{eq:mpc}. Moreover, the controller would be effectively blind to unanticipated dangers while executing the sequence: by the time $T$ time steps are up, it may be too late to bring the system back to safety.

Thus, the second key idea of MPC is to resist the temptation to use the full input sequence:
an MPC controller solves~\eqref{eq:mpc} at each time step, \emph{and applies only the first computed input}. Using only the first control action is how MPC mitigates potential dangers while retaining optimal behavior.

Model predictive control (MPC) has been widely used to achieve anticipatory control of HVAC system for energy saving while ensuring human comfort~\cite{DRGONA2020190}. MPC-based control methods require the solving of a multi-step optimization problem so as to achieve the control sequence. The control output at the current stage is exerted and this process is repeated with the evolving of time. MPC-based control methods are online control methods that rely on the dynamic predictions of system disturbance. For HVAC control, several critical issues are related to the development of MPC-based controller: modeling complexity, modeling accuracy and prediction accuracy. MPC-based controllers have been popular with HVAC control (see \cite{DRGONA2020190, afram2014theory, kathirgamanathan2021data, oldewurtel2012use} for a comprehensive review).The MPC controllers generally use either physics-based models (also known as ``analytical first principle'' or ``forward models'') or data-driven models (also known as ``black box'' or ``inverse models'') to predict system output. 

MPC-based controllers have many advantages over rule-based or PID controllers. Since the energy saving and human comfort targets may be directly expressed in the MPC's objective and constraints and the time-varying disturbances are directly considered, MPC shows superior performance in energy efficiency and in ensuring human comfort. In addition, MPC-based controllers are robust to the time-varying disturbance, quick transient response to the environment changes, and consistent performance under a wide range of varying operating conditions.  


However, MPC's advantages are achieved at the cost of frequently solving comprehensive optimization problems on-line. The intrinsic problem related to the HVAC control is non-linear and non-convex.  This leads to the high online computation cost. Many efforts have been made to address the computational challenges. One solution is to restrict attention to \emph{linear models} of the form $x(t+1)=A(t)x+B(t)u$, where $A(t)$ and $B(t)$ matrix functions which whose values may be time-dependent but not state-dependent. This model simplification, in conjunction with a quadratic objective, reduces~\eqref{eq:mpc} to a \emph{quadratic program}, a convex problem that even embedded devices can solve quickly. However, a simpler model will necessarily lead to predictions of lower accuracy, and thereby worse performance. If we do not wish to simplify the model, an alternative is to develop explicit MPC controllers \cite{parisio2014control, klauvco2014explicit, Drgona2013}.  The key idea is to drive some close-form solution for the MPC problem and then identify the parameterized control outputs on-line. To achieve the objective, simplified or approximated linear models for HVAC control are generally used. The benefit of explicit MPC is that it can be implemented on simple hardware and yields low on-line computation cost. Another line of work to address the computation burden is to develop distributed solvers. When MPC-based controllers are applied to large-scale commercial buildings, we require to solve large-scale optimization problem that couples the control of the HVAC system supplying multiple zones or rooms. Centralized methods tend to suffer from the high computation cost and are not scalable.  Many works have made effort to develop distributed MPC controllers \cite{yang2020hvac, yang2021distributed, zhang2017decentralized, radhakrishnan2016token, radhakrishnan2017learning}.  The major challenge to develop distributed MPC controllers for HVAC system  is to handling the couplings across the multiple zones, including the thermal couplings (i.e., heat transfer), the composite objective (i.e., the fan power is quadratic w.r.t. the total zone mass flow rates), and operating constraints).  The optimization problem related to multi-zone HVAC system is a nonlinear and nonconvex optimization problem subject to nonlinear couplings (nonlinear and nonconvex). The existing distributed solution methods can not be directly applied to solve such a class of problems. To overcome the challenges and enable the decomposition of the problem, \cite{zhang2017decentralized, yang2020hvac, yang2021distributed} has explored the convex relaxation technique to solve such problem.  Particularly, \cite{yang2020hvac, yang2021distributed} have applied the well-known alternating direction method of multipliers (ADMM) to solve the resulting relaxed problems which have nonconvex objectives but subject to linear couplings. The superior performance both in energy saving and computation efficiency have been demonstrated via simulations. Slightly different, \cite{radhakrishnan2016token, radhakrishnan2017learning} have explored the decomposition of the problem w.r.t. the zone controller via hierarchical optimization and relaxation. Particularly, these works have relied on predictions to predict the thermal couplings across the zones. Besides, the objective components are optimized in a two-step. 
These works have comprehensively discussed the challenges to develop distributed control methods for large-scale commercial buildings due to the problem complexity.

\subsubsection{Reinforcement Learning}

Time-varying and uncertain parameters such as the weather variations and occupancy represent one major challenge for HVAC control. 
MPC controllers have relied on short-term dynamic predictions to address such problems but at the cost of high on-line computation burden and an accurate model to predict the system state evolution.  These are  the two main obstacles for the wide deployment of MPC controllers in practice.  As one main branch of ML, reinforcement learning (RL) explores  to  optimize the operation of system through the interaction with  environment as long as the performance can be observed and quantified.  

RL being one of the mainstream tools for multi-stage decision making under uncertainties \cite{sutton2018reinforcement} makes it well suited for HVAC control.  Indeed, RL has been widely used for advanced control of HVAC systems considering the uncertainties and multi-stage problem features \cite{yang2021stochastic, sun2012building, sun2015event}.  RL can enable efficient on-line computation by learning the control policies off-line.  Specifically, by learning the control policies off-line,  the on-line implementation of RL only requires to search the table to identify the mapping from the state space to the action space.  

However, the obstacle to effective implementation of RL is that the learning may be computational intensive or intractable when state and action space are large, either in terms of dimensionality (number of observations) or enumeration of action space (continuous vs. categorical.) To overcome the challenges, deep reinforcement learning (DRL) has been popular in this domain. DRL proposes to combine the approximation and characterization capabilities of deep neural networks with RL techniques. Specially, deep neural networks are used to characterize the state or state action value functions. The benefits of DRL include reducing the quantity of data required for training and reducing the memory required to store the policy for on-line implementation.  The proliferation of smart meters have boosted the development of DRL-based HVAC controllers for smart buildings. We refer the readers to \cite{yu2021review, han2019review, mason2019review, wang2020reinforcement, yang2020reinforcement}.  As an earlier work, \cite{wei2017deep} has explored deep Q-network based RL for single zone HVAC control.  An Q-network is trained to approximate the state-value functions to overcome the computation burden related to the large state space. \cite{yu2019deep} has explored the application of Deep Deterministic Policy Gradients (DDPG)  \cite{lillicrap2015continuous} for the management of building energy system including an HVAC system.  The benefit of DDPG is that it  can enable the decision making on the continuous state and action space so as to improve the control accuracy.  While most of the existing works have focus on the control of HVAC system for single zone or single room, several recent works have explored the development of distributed  RL for multi-zone commercial buildings \cite{yu2020multi, hanumaiah2021distributed}. For example,  \cite{yu2020multi} studied the application of multi-agent actor-attention-critic RL \cite{iqbal2019actor} for multi-zone HVAC control considering both thermal comfort and indoor air quality. 

One major advantage of RL based HVAC controllers over MPC is that it does not depend on explicit and accurate models, as in classical RL methods, a model of the environment may only ever be implicitly encoded in the inner layers of the policy or value prediction.  In this way, RL methods alleviate the burden on the engineer for articulating an accurate model of the environment. It also allows for environment adaptation, i.e. flexibility to an environment that changes over time. The RL agent would just shift the value prediction it makes from actions over time. 

RL methods are generally off-line tools and require a rich set of data to compute the optimal control policies for a wide range of possible operation conditions.  Moreover, the existing RL models including DRLs are still suffering from the computation intensity and scalability issues. Classical RL works best in cases where failure, at least at first, is relatively cheap. However, numerous work elucidate ways that RL can be modified to reduce variability and increase safety. Arnold et al use Surprise Minimizing RL, a type of RL that rewards the agent for achieving states closer to those seen before, in optimizing building demand response \cite{arnoldSMIRL}. Augmenting reinforcement learning with a system simulation is important, as almost always a large part of the training can take place in a simulation environment which can use a rules-based heuristic as a starting point, and use exogenous parameters to create a distribution of unique systems to train on (i.e. domain randomization). Finn firmly established the field of meta-learning in RL, arguing that a technique like Model Agnostic Meta-Learning can train in the different simulations to approximate a starting distribution for policy network weight initializations \cite{FinnMAML}. Jang et al. studies offline learning through offline-online RL \cite{jangOfflineOnline} and Model Agnostic Meta Learning (MAML) \cite{jangMAML} to get RL agents ``hit the ground ready'' using offline datasets and simulations. 

\subsubsection{Learning-based Model Predictive Control}
From the above literature, we note that MPC and RL tools both have pros and cons for HVAC control.  MPC-based controllers can yield superior performance both in energy saving and ensuring human comfort due to the capability to incorporate the dynamic predictions regarding the disturbances. However, this is achieved  at the cost of high on-line computation burden.  Moreover, MPC-based controllers generally suffer from the modeling complexity.  For example, the more elaborate thermal comfort model, such as the Predicted Mean Vote (PMV) is hard to be incorporated in the MPC-based control due the challenge to solve the optimization problem \cite{xu2017pmv}.  In contrast, RL based models can accommodate the modeling complexities but require a huge amount of data to ensure the consistence of performance under uncertainties.   The current research trends has been the combination of MPC and RL so as to enjoy both of their pros. The researches are mainly in the following two aspects.  One is to use ML to mimic the control policies output by MPC controllers.  Specifically, the control of MPC controllers can be regarded as the parameterized policy with respect to the disturbance. To reduce the on-line computation burden and enable the deployment of MPC controller on simple hardware, the characterization capacity of ML, such as deep neutral networks can be leverage to  train parameterized control policies. This can yield control policies of high energy saving performance and human comfort with MPC at lower on-line implementation cost.  For example, \cite{drgovna2018approximate}  proposed an approximate MPC where deep time delay neural networks (TDNN) and regression tree based regression models are used to mimic the control policy of MPC controller.  Simulation results show that the approximate MPC only incurs a slight 3\% of performance loss but reduces the on-line computation burden by a factor of 7.  Later,  a real-time implementation of the  approximate MPC is  developed  and demonstrated in \cite{drgona2019stripping} in an office building located in Hasselt, Belgium.  Similarly, \cite{karg2018deep} proposed to use  deep learning networks to learn the control policies of MPC controller to enable low-cost implementation.  Using ML tools to learn the control policies of MPC controllers can be regarded as the extension of the explicit MPC discussed above. When the problem is simple and the closed-form solution is available, the implementation of explicit MPC only requires to compute the parameters of the parameterized policy and therefore the on-line implementation only requires low on-line computation cost.  By leveraging ML tools, we are able to achieve the same objective by training the parameterized policies using deep neural networks when the problem is complex and the explicit closed-form solution is not accessible.  Another line of work to combine ML with MPC is to use ML to establish the models for some complex components with building or HVAC system.  We refer the readers to \cite{afram2017artificial} for a comprehensive review.  For example, neural networks can be trained to efficiently predict PMV~\cite{ku2014automatic, ferreira2012neural} for a MPC controller.  Since neural networks are generally characterized by nonlinear functions,  the nonlinear solvers, such as genetic algorithms and particle swarm algorithms are often used to solve the resulting optimization problem. Compared with the original PMV models, neural network approximation can enable more efficient evaluation of the PMV index without requiring any iterations.

\subsubsection{Summary}
Though MPC and RL are two main tools for HVAC control, there has not been significant implementation of them in buildings, where the primary controllers have been PID and rule based ones. There are still many practical issues required to be addressed to enable the deployment of MPC and RL based methods. 
Challenges related to classic MPC controllers include the development of models that can properly trade off the modeling complexity and accuracy.  In addition, the performance of MPC-based controllers is closely related to the prediction accuracy of disturbance and the prediction horizon. Last but not the least, MPC based controllers suffer from the high on-line computation cost and  the implementation requires advanced micro-controllers.   In contrast, RL based controllers are free from the establishment of models and can enable efficient and low-cost implementation but requires enormous data to train the agent.  Moreover,  RL based controllers can not incorporate the possibly available dynamic predictions to enhance the performance. That's why most of the existing RL based controllers are not comparable with MPC based controllers in terms of energy saving performance.  Another critical issue related to MPC controller is that the safely can not be ensured. Since the agent is trained on limited data, RL based controller may draw the system to any possible state which may be prohibitive in practice.  Combing ML with MPC-based controllers seems a promising solution as it can enjoy  both the pros of ML and MPC.  However,  how well the control policies can be learned  and how the network should be designed remain to be investigated. 
The operation of HVAC system is expected to provide a  comfortable indoor environment for the occupants.

\subsubsection{Mechanism Design}

Compliance is the ability for people, in aggregate, to follow guidelines or recommendations. In any system that seeks to influence human behavior, lack of compliance is generally a large problem. Energy systems are no different: an empirical study of demand response found that non-compliance weakened the overall effect of the programs\cite{CAPPERS20101526}.

\begin{figure}
    \centering
    \includegraphics[width=\columnwidth]{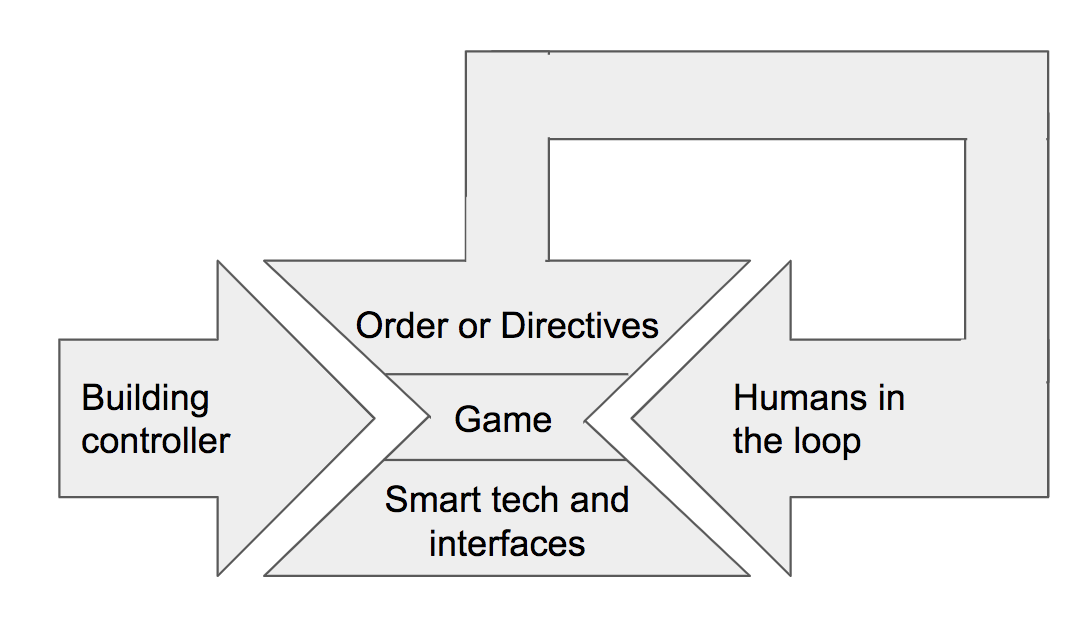}
    \caption{An example of a system in which gamification can be part of a control strategy.}
    \label{fig:game}
\end{figure}

Gamification of systems, i.e. the adding of an additional layer of intrinsic motivation to a core program, may be an important tool to boost compliance. The additional layer of motivation may come in the form of engagement-derived pleasure (i.e. interesting and exciting gameplay), competition (i.e. equally matched competitors), or social status (i.e. badges, ranks, and advancement.) Hamari et al. \cite{gamification_review} rightly note the explosion in interest in gamification, from gamification around code learning to gamification around energy demand reduction and perform an extensive review of its efficacy, finding that generally gamification provides positive results. For a visualization of the purpose that gamifying provides, please see Figure \ref{fig:game}.

There are some examples of gamifying office-related energy systems in the literature. Although not extensively, ML algorithms have been utilized in the gamification process to improve some individual components of the game. In one example, an game called ``NTU Social Game'' has been formulated around reductions in energy in the a residential dormitory. Here, players compete against each other to have the lowest energy, and accrue points depending on how much less energy they consume. Prizes are not simply doled out according to whoever has the highest energy reductions; instead, players are entered into a prize pool if they are among the top third performing participants. A small number of prizes are allocated randomly amongst the winning pool as per the Vickrey Clarke Groves (VCG) auction mechanism. A design like this is shown to boost user engagement and directive compliance \cite{konstantakopoulos2019deep,konstantakopoulos2019design}. The NTU social game has been used for energy reduction and to test different visualizations on player outcome \cite{spangherVisualization}. It has also been used to conduct segmentation analysis of players into low, medium and high energy-efficient groups for intelligent incentive planning~\cite{das2019novel,das2020occupants}. The authors use a hybrid approach of supervised classification and unsupervised clustering, to create the characteristic clusters that satisfy the energy efficiency level. They also have incorporated explainability in the model by combining both the ML approaches.

An extension of this work, ``OfficeLearn'', considers prices that are distributed across a day. Here, a building controller sets energy prices throughout the day, and players compete with each other on having the lowest cost of energy throughout the day. As before, the top performing individuals are entered into a prize pool and selected based on a VCG auction mechanism \cite{spangher2020prospective}. This paper presents
an OpenAI gym environment for
testing demand response with occupant level building dynamics, useful for standardizing RL algorithms implemented for the same purpose.

Other examples of energy based games exist. In one game called ``energy chickens'', competitors save energy from smart appliances for the health of virtual chickens; depending on how healthy the chickens are, they will generate probabilistically more or less eggs \cite{orland2014saving}. In another game, ``Energy Battle'', players in simulations of identical residential homes strive to reduce energy and players compete one to one without a VCG mechanism \cite{geelen2012exploring}. In a third game, ``EnerGAware'', players also compete in a building simulation in which players, who are virtual cats, decide how to operate a house's appliances with greater energy efficiency while not reducing any overall functions \cite{casals2020assessing}. The use of ML in energy social games, although not prominent before, is showing signs of increase, and is adequately justified since ML can have a profound use in coordinating multi-agent systems such as a game.

\subsection{Integration of Buildings with Power-Grids, Microgrids and Energy Markets}

So far, we have discussed how modeling and control methods can be used to improve occupant comfort and increase the energy efficiency of buildings. However, buildings do not function in isolation and their operation can actually have significant impacts on other energy systems. The energy utilized in buildings come from power grids, and recently, from renewable sources installed at the local building/community level as well. In order to mitigate climate change, buildings must maximize their energy consumption from clean sources. 

One way that buildings can achieve the above is by providing load flexibility to facilitate clean energy sources such as wind and solar that have variable generation and can not be controlled at will to meet electricity demand, a mechanism commonly called Demand Response (DR). By making their electricity consumption flexible, buildings can adapt to variable generation curves and enable greater amounts of clean generation on the power grid. 
Buildings can capitalize on their load flexibility using two methods: by participating directly in demand response programs run by utilities, or by forming aggregations and participating in broader energy markets.

Another way that smart buildings can mitigate carbon emissions is by investing in local clean generation resources (like solar). Smart buildings are often equipped with distributed energy resources (DERs) such as rooftop solar, backup batteries, and electric vehicles (and chargers) to reduce demand charges or optimize grid consumption. If these resources are associated with a single building, they may not be fully utilized. Therefore, buildings should function as \emph{prosumers}, i.e. use their electricity production capability to provide services to other buildings or to the power grid. Doing so can increase the utilization of clean energy sources, and incentivize additional investment in DERs thus reducing the reliance on grid-wide CO$_2$ emission. In the following parts, we will discuss demand response in further details, and how machine learning plays a role in achieving the same. We will also cover on the prosumer nature of buildings.

\subsubsection{Demand Response (DR)}
We start off by discussing how individual buildings can use their load flexibility as demand response resources.
Demand response is a broad term for a variety of methods used to module electricity demand. These can include load response to time-varying electricity rates, financial incentives, or direct control of loads (like appliances and air conditioning) by utilities. 

\paragraph{DR sources and timescales}

There are two primary sources of flexible loads in smart buildings: lighting and the HVAC system \cite{watson2006strategies}. Modifying energy consumption for either of these has a direct impact on the building occupants, as covered in the thermal comfort section (Sec.\ref{sec:thermal-comfort}). Lighting control is a great DR resource since it does not have a \emph{rebound} effect, i.e. curtailing lighting load during a DR event does not cause a spike after the DR period ends.

Flexible load can be modulated over timescales ranging from a few seconds to a few hours \cite{alstone20172025}. For example, smart buildings can \emph{shift} load by pre-heating or pre-cooling their indoor spaces, and can \emph{shed} load altogether by reducing their lighting levels. They can also modulate their instantaneous load by adjusting fan speeds, and respond to high frequency regulation signals in electricity markets.

\paragraph{DR program mechanisms}
The two most common types of demand response schemes are time-of-use (ToU) rates and curtailment contracts. Time-based rates aim to \emph{shape} the energy consumption curve by driving long-term behavioral change and energy efficiency investments \cite{alstone20172025}. Utilities also run DR programs to procure short-term load curtailment, e.g. in situations when the electricity price spikes and it is too expensive to procure additional generation. These programs are typically structured as contracts where the buildings promise to curtail their energy consumption by a certain amount when called upon to do so, and the utility pays an incentive upfront based on the curtailment size. If the building is unable to curtail its consumption by that amount, it has to pay a penalty to the utility proportional to the shortfall. Readers are encouraged to refer to \cite{albadi2007demand} for an overview of common demand response mechanisms.

\paragraph{ML for forecasting loads and baselines}
Buildings procure  energy from  utilities, who buy it from from generators through long-term contracts or in wholesale energy markets. Energy consumption is variable, and utilities employ many machine learning methods to forecast electricity consumption accurately. Authors in \cite{energyprimer} provide an overview of the functioning of electricity markets. The electricity consumption forecasts are usually made for the aggregate load, and individual building loads are not as important for utilities. However, there are a couple of use cases for load forecasting at the building level. 

First, in order to implement DR programs and incentivize loads to modulate their energy consumption, it is important to estimate the counterfactual energy consumption: what would the load have consumed without an incentive/intervention? This is considered as the baseline, and needs to be developed at the load level, i.e. building level. There are a number of ML methods used for developing baselines. \cite{deb2017review} conduct a review of time series models for forecasting building energy consumption. They found that a variety of ML models have been used for this purpose: classical time series methods such as ARIMA, support vector machine (SVM) regression for short-term and long-term forecasts, as well as neural network based methods. \cite{jazaeri2016baseline} compare different methods for computing baselines, such as regression, neural networks, polynomial interpolation and classical methods (X of last Y days) used by utilities. They find that the machine learning methods emerge as the state-of-the-art for most of the cases.

A second use case of forecasting load at the building level is peak shaving, where buildings aim to forecast and then reduce their peak energy consumption in order to reduce the demand charges that they have to pay as a part of their electricity bill. \cite{xu2019probabilistic} develop a probabilistic forecasting methodology to predict the time of occurrence of magnitude of the peak load, where they quantify the probabilistic occurrence and magnitude of the peak load. \cite{kim2019short} use a variety of time series methods to forecast the peak load demand of institutional buildings, with a dataset of buildings in South Korea.

\paragraph{ML for estimating DR capability}

Once the baseline energy consumption is estimated, the utility/system operator can compare it with the actual consumption to determine the impact of the DR program, and in turn determine payments and penalties. However, most energy consumption by buildings is variable, and depends on occupancy and weather conditions. Additionally, the impact of load reduction measures such as reducing lighting and changing temperature setpoints are difficult to estimate ahead of time since the amount of energy curtailed will depend on the building state before the intervention. Some work has been done to quantify the effect of DR actions, e.g. \cite{nghiem2017data} develop a Gaussian process model of how buildings respond to DR signals.  In DR contracts, there might be a penalty associated with curtailment shortfall, and accurately estimating curtailment capability becomes an important task. \cite{mathieu2011quantifying} use linear regression to develop models for estimating demand response capability using time-of-week and outdoor air temperature buckets. It is important to have such DR capability estimates over longer time horizons, and \cite{jung2014data} develop a look-up table approach to predict future DR capability. They relate power consumption data to parameters such as time-of-day, day-of-week, occupancy, outdoor temperatures and determine the load flexibility at a point when the forecast value of these parameters is available.


\subsubsection{DR Aggregations}
Load flexibility can be procured through utility programs as discussed previously, and also through direct market participation by loads. There has been a move in recent years to allow loads to submit bids for load reduction into wholesale electricity markets, and treat those bids similar to generation bids. Federal regulations, specifically Federal Energy Regulatory Commission (FERC) Order 2222 in the United States have mandated that distributed resources be allowed to participate in wholesale markets. The state of California has instituted a market mechanism called Demand Response Auction Mechanism to allow loads to submit bids as `negative generation'. However, the magnitude of load flexibility provided by individual buildings is often too small for them to participate directly in electricity markets. This is where aggregations can play a role: aggregators can bundle the load modulation capability of multiple buildings, and use the aggregate capacity to submit bids to electricity markets. Demand response can also be coupled with other resources like generation and storage, and form a part of a \emph{virtual power plant}.

\paragraph{Value proposition of aggregations}
Aggregations can add value beyond simply providing a means of market participation, by spreading the risk of shortfall of curtailment capability. \cite{burger2016} discuss the value of aggregators in electricity systems, and identify that apart from opportunistic and transitory value, aggregations also provide fundamental value, e.g. through economies of scale, that is independent of market or regulatory context. Since the load curtailment capability of buildings is variable, and each building faces a risk of having to pay penalties if it curtails less power than it promised to. However, different buildings will have different variability in their load curtailment capability, and aggregations can capitalize on this uncorrelatedness to minimize the penalty risk for all the participants \cite{agwan2021asset}. Optimizing the complementary nature of buildings can significantly increase the marginal value of forming an aggregation. 

\paragraph{Optimizing market bids}
Once an aggregation is formed, it can optimize its bid in electricity markets using price forecasts and predictions of the available curtailment capacity. A variety of time series methods are used by commercial entities that participate in these markets, and improving on these methods is an active area of research in both academia and industry as it can directly lead to higher revenues. \cite{shinde2019literature} conduct a review of electricity price forecasting methods. \cite{brusaferri2019bayesian} develop probabilistic forecasting methods for electricity prices using Bayesian deep learning methods, as opposed to commonly used point forecasts.

\subsubsection{Prosumers and Local Energy Markets}

So far, we covered how buildings can use their load flexibility to provide services to the power grid, both individually and through aggregations. Additionally, buildings can directly consume energy from onsite clean resources. 
Smart buildings are often equipped with a variety of generation and storage resources such as rooftop solar, battery backups and electric vehicle chargers which can be used to reduce demand charges or optimize grid consumption. They can profit from these resources by supplying energy back to the grid, i.e. functioning as \emph{prosumers}. A prosumer is an entity that is capable of energy production as well as consumption through the presence of local generation and energy storage devices. 

\paragraph{Value Proposition of Prosumers}
Investing in local energy resources is not always a profitable proposition: buildings have to invest in over-capacity generation to meet peak loads, and storage might only be used for peak shaving on a few days in a month or a few hours in a day. If these resources can be shared across multiple users, their utilization levels can increase which can further incentivize investment in clean energy sources. One way to facilitate such sharing is through local energy markets, where prosumers can trade energy with each other before procuring the aggregate balance from an outside source, such as the utility \cite{agwan2020optimal}. 

\paragraph{Organization of Local Energy Markets}
There are two main challenges that come up while organizing local energy markets: selecting prosumers to form a market, and then managing the trades within the market. Machine learning methods are used in tackling both of these.

First, we consider the problem of setting up a local energy market, and the methods used to address this challenge. The authors in \cite{saad2011coalitional} use coalitional game theory to optimally aggregate prosumers in a cooperative microgrid. \cite{wu2020stochastic} uses a mixed-integer linear programming to discuss the optimal sizing of energy resources in a microgrid. In \cite{quashie2018optimal}, the microgrid resource planning problem is cast as the upper level of a bilevel program which aims to optimize investment costs along with system reliability. Research in this area mainly focuses on optimizing resource investments, and the problem of evaluating a new prosumer using numerical metrics is not as widely studied. 
While forming a local energy market, it is valuable to select participants that complement each other's net load curves. For example, a building that has a net load at a certain time would benefit from being in a market with a net generator at that time. The complementarity of market participants is essentially a measure of how well the electricity generation and demand curves match up. There have been game theoretic approaches to evaluating complementary participants, and the authors  in \cite{agwan2021optimal} have developed a complementarity metric that uses the net generation and load curves to evaluate the marginal value of adding new participants to the local energy markets.

Once a market has been set up, we need to facilitate energy trades between participants. This can be achieved by soliciting demand and supply bids from participants, and then settling the market. However, it may be an onerous task for buildings to generate such bids for a market. An alternative to this approach is to set a price for energy and then settle all trades at that prices. A number of previous works develop price based controls: \cite{kim2016online} develop a price to minimize load variation, and \cite{liu2017energy} develop a price based on supply-demand ratio to incentivize energy sharing.

While this price can be set in an iterative manner using a variety of distributed optimization techniques, it might be preferable to set it in a one-shot manner to avoid the time lag of back-and-forth communications. This method has been investigated in literature, e.g.
\cite{jia2013retail} develop an online learning method to dynamically price electricity for retail consumers. They try to learn the response function of a number of price-sensitive consumers who modify their HVAC load to optimize their consumption. \cite{agwan2021pricing} approaches this problem using reinforcement learning, and develop a controller that can set day ahead electricity prices for a local market with a variety of prosumers.

Summarizing, ML can aid in predicting the energy load of buildings, enable demand response schemes via load flexibility, and can accelerate the energy market set up with buildings as prosumers.

\section{Limitations of Machine Learning in the context of Smart Buildings}

Although machine learning techniques have proliferated across many fields to great success, the lack of success of machine learning in some applications bears note. There are some notable applications where ML doesn’t work well: safety critical applications, data poor applications; we describe the fundamental barriers to success in these domains. Arguably, most of these barriers also exist when ML is employed in smart buildings. 

\subsection{51/49 vs. 99/01 Problems} Does the area of interest require 51\% and 49\% failure of the control algorithm in its trials, or does it require 99\% success and 1\% failure? Examples of the former may include stock investing, product recommendation, efficient allocation of computational tasks, and image classification. Examples of the latter may include medical decisions, decisions in autonomous driving vehicles etc.

In smart buildings, the applications fall into both the above categories, and it often becomes harder for ML algorithms to achieve 99/01 performance without jeopardizing robustness and safety. Advancements in machine learning that may help to address 99/01 problems in buildings are safety-guarantees in machine learning, robust worst case minimizing reinforcement learning, proper explainability etc.

\subsection{Data Problems} Does the domain provide rich, reliable, and accurate data? The shining examples of machine learning success often occur in domains where this is the case: reinforcement learning techniques started to shine when Atari game simulators were developed and could be run millions of times, image ML began to flourish when large image datasets were provided and standardized, and Natural Language Processing began to bear fruit when datasets like the entirety of Wikipedia could be processed and trained on. Oftentimes, building energy datasets are not as large, rich, or diverse, and so many of the techniques that work well in different domains may not translate well to energy~\cite{livcina2018development,dong2022global,fierro2019mortar,miller2020building}. A fruitful area of work may include preparing and standardizing large datasets for ease of energy research. Advancements in RL that can work well in data poor regimes may include generative models, regularized and sparse models etc.

\subsection{Pathway to Physical Implementation} Does the domain have a physical interface that can use the output of machine learning models? Oftentimes we may put a lot of work into making a prediction model accurate, but a building from the 70’s might not have the physical infrastructure to house the model, respond to its predictions, or reliably provide state information. A smart appliance may be limited in its scope and effect. To this extent, building Operating Systems (OS) are newer developments that seek to unify buildings into one software control scheme. Another important function of building OSes should be to provide to the outside world an API that does not need to be aware of inner complexities of the building, but can provide certain data about the building (aggregate energy usage) and receive certain grid information like price and state. A third important function of the building OS is to provide unified controls. To this extent, machine learning research in controls is inherently closer to many of the applications that a building may need. 
\section{Conclusion and Future Work}~\label{sec:future_works}
During the life cycle of a building including the design, construction, operation and maintenance, retrofitting, and disposal, there are many factors that contribute toward a building's energy consumption, efficiency, and embodied carbon. 
In this paper, we primarily evaluated improving energy efficiency in buildings via ML during design, operation, and maintenance stage. 
From the well-being of an individual, to building's occupants, to the building operations and maintenance, and finally to the electric grid impact and energy market economic response, we discussed all the different facets of the smart building ecosystem.

While ML shows promising results in facilitating the energy improvement of building operation and maintenance, there are still several challenges and works to alleviate them ahead. 
The foremost challenge of all ML approaches is the time and effort to collect reliable data. 
ML models require a great amount of data, and the quality and the selection of the data greatly affects the quality of the output model.
In thermal comfort modeling, data is even harder to come by as it uses intrusive methods that require occupants' input.
Additionally, thermal comfort data is inherently class-imbalanced, which may result in ill-conditioned models.
In building modeling, realistic data such as occupancy, temperature settings, and schedules are crucial, and without them, the outcomes of models may be inaccurate or have a high level of uncertainty.
Furthermore, the realistic data required may be completely dependent on the application in which the data is needed.
Although many researchers have been collecting this realistic data, there is no consistency between companies or research entities.
As such, utilizing the collected data across many applications is time consuming and ultimately infeasible.
A potential solution for the above data issue is to develop a standard for collecting and storing building data, potentially as a public data repository from which crowd-sourced data can be posted and requested.
An example of a data standard that could suffice is the BRICK schema \cite{balaji2018brick}, which is a metadata schema that represents buildings' sensor information and the relationships between them. Grey-box models developed on the data along with building physics can be the ultimate goal. Synthetic data generation methods are also being proposed as a solution to the data insufficiency problem~\cite{das2022conditional}.

Generalization is another open challenge for building applications. 
Firstly, thermal physiology and expectations vary significantly between human occupants.
If all occupants' thermal comfort models need to be individually trained, it will be computationally expensive. 
As a result, the need for more generalized models that can represent all occupants arises. 
Secondly, as all buildings are inherently different in design and operation, models and control strategies from one building are difficult to apply to other buildings. 
Hence, transfer learning and domain adaptation are some open areas of research that help create more generalized building models. 

Finally, thermal comfort models, building models, control algorithms, and grid applications exist on different platforms and do not have a standard interface through which they can all be used.
Developing a standardized platform for each application and a standardize interface protocol for the above information can benefit the research communities, as the application of the above techniques are correlated. 

As data becomes more widely available through IoT and ML becomes more computationally feasible, more and more success can be achieved in enhancing occupant comfort and building energy performance. 
In the future, ML could provide resiliency in the built environment and assist in optimizing occupants' satisfaction, energy use and cost.

\section{Acknowledgements}

This research is funded by the Republic of Singapore's National Research Foundation through a grant to the Berkeley Education Alliance for Research in Singapore (BEARS) for the Singapore-Berkeley Building Efficiency and Sustainability in the Tropics (SinBerBEST) Program. BEARS has been established by the University of California, Berkeley as a center for intellectual excellence in research and education in Singapore.
\section{Declaration of Competing Interest}
The authors declare no sources of competing interest(s).

\bibliographystyle{elsarticle-num}
\bibliography{references}

\end{document}